\documentclass{article}

\usepackage{PRIMEarxiv}

\usepackage[utf8]{inputenc} 
\usepackage[T1]{fontenc}    
\usepackage{hyperref}       
\usepackage{url}            
\usepackage{booktabs}       
\usepackage{amsfonts}       
\usepackage{nicefrac}       
\usepackage{microtype}      
\usepackage{lipsum}
\usepackage{graphicx}       
\usepackage{amsmath}        
\usepackage{siunitx}        
\usepackage{float}

\title{
Sequential Reservoir Computing for Efficient High-Dimensional Spatiotemporal Forecasting
\thanks{\textit{\underline{Citation}}: 
\textbf{Akbari Asanjan, A., Wudarski, F., O'Connor, D., Geaney, S., Strbac, E., Lott, P. A., \& Venturelli, D. Sequential Reservoir Computing for Efficient High-Dimensional Spatiotemporal Forecasting arXiv preprint arXiv:2301.00001.}}
}
\author{
    Ata Akbari Asanjan, Filip Wudarski \\
    USRA Research Institute for Advanced Computer Science (RIACS) \\
    California \\
    \texttt{\{aakbariasanjan, fwudarski\}@usra.edu} \\
    \And
    Daniel O'Connor, Shaun Geaney, Elena Strbac \\
    Standard Chartered Bank, 1 Basinghall Avenue \\
    London, UK \\
    \texttt{\{daniel.oconnor, shaun.geaney, elena.strbac\}@sc.com} \\
    \And
    P. Aaron Lott \\
    USRA Research Institute for Advanced Computer Science (RIACS) \\
    California \\
    \texttt{plott@usra.edu} \\
    \And
    Davide Venturelli \\
    USRA Research Institute for Advanced Computer Science (RIACS) \\
    California \\
    \texttt{dventurelli@usra.edu} \\
}

\begin{document}
\maketitle

\begin{abstract}
Forecasting high-dimensional spatiotemporal systems remains computationally challenging for recurrent neural networks (RNNs) and long short-term memory (LSTM) models due to gradient-based training and memory bottlenecks. Reservoir Computing (RC) mitigates these challenges by replacing backpropagation with fixed recurrent layers and a convex readout optimization, yet conventional RC architectures still scale poorly with input dimensionality.
We introduce a Sequential Reservoir Computing (Sequential RC) architecture that decomposes a large reservoir into a series of smaller, interconnected reservoirs. This design reduces memory and computational costs while preserving long-term temporal dependencies. Using both low-dimensional chaotic systems (Lorenz63) and high-dimensional physical simulations (2D vorticity and shallow-water equations), Sequential RC achieves 15–25\% longer valid forecast horizons, 20–30\% lower error metrics (SSIM, RMSE), and up to three orders of magnitude lower training cost compared to LSTM and standard RNN baselines.
The results demonstrate that Sequential RC maintains the simplicity and efficiency of conventional RC while achieving superior scalability for high-dimensional dynamical systems. This approach provides a practical path toward real-time, energy-efficient forecasting in scientific and engineering applications.
\end{abstract}




\section{Introduction}
In recent years, the field of scientific research has undergone a remarkable transformation propelled by the swift advancements in machine learning techniques. These progressions have ushered us into a new era of possibilities for understanding and predicting complex phenomena across a wide range of research areas \cite{vaswani2017attention, gers2000learning, he2016deep}. Notably, the prevailing trend in enhancing these cutting-edge models revolves around augmenting the complexity of their architectures. This often translates to an escalation in the number of parameters, a key aspect that influences the model's capacity to capture intricate patterns and relationships within the data \cite{lawrence1998size, golubeva2020wider, gers2000learning, vaswani2017attention}. Advanced architectures exhibit remarkable capabilities in handling complex datasets and extracting nuanced patterns that were once challenging for traditional approaches. However, a noteworthy consequence of this architectural sophistication is the increased demand for computational resources, particularly in terms of memory and processing power \cite{desislavov2023trends}. As a result, balancing the quest for sophisticated architectures with considerations for computational efficiency becomes paramount in ensuring that the scientific community can harness the full potential of machine learning advancements across various disciplines.

One notable illustration of the progression toward more sophisticated architectures in time-series analysis is the evolution from Elman-type Recurrent Neural Networks (RNNs) \cite{chung2014empirical} to the advent of gated structures such as Long Short-Term Memory (LSTM) \cite{gers2000learning} and Gated Recurrent Unit (GRU) \cite{cho2014properties}. This architectural refinement in recurrent models has proven instrumental in extending their capacity to comprehend longer sequences and mitigating training challenges associated with suboptimal learning schemes, such as the vanishing and explosion of gradients \cite{mikhaeil2022difficulty, akbari2018short, chung2014empirical}.

The introduction of gated recurrent models has effectively addressed learning issues by incorporating gate units that regulate the flow of information into the core cell of the recurrent model. This mechanism enables the preservation of crucial information while filtering out irrelevant details \cite{gers2000learning, chung2014empirical}. This enhancement, however, comes at the cost of a substantial increase in the number of parameters—roughly three to four times more in each recurrent layer \cite{salehinejad2017recent}. Despite this trade-off, the improved ability of gated recurrent models to overcome inherent training difficulties has positioned them as pivotal tools in time-series analysis, demonstrating the ongoing trend of balancing model sophistication with the pragmatic considerations of computational efficiency and training effectiveness.

Reservoir Computing (RC) has undergone a resurgence in recent years, gaining renewed attention for its unique merits of efficient training and accurate prediction. The genesis of RC traces back to seminal work \cite{buonomano1995temporal} where the concept was conceived. This pioneering effort introduced a network of random and fixed spiking neurons configured as dynamic synapses, coupled with a trainable output layer employing a correlation-based training approach. The inspiration derived from this work laid the groundwork for the development of two major RC frameworks: Jaeger's Echo State Network \cite{jaeger2001echo} and Maas' Liquid State Machine \cite{maass2002real}. 

In Echo State Networks (ESNs), the mainly-known architecture for reservoir computing, the author delineates the echo state properties, asserting that the model's state should ultimately hinge solely on the driving input signal. In other words, the state becomes an echo of the input, and the impact of initial conditions should gradually diminish over time \cite{jaeger2001echo}. This property is realized via a composition of an input layer and a reservoir layer comprising recurrently connected neurons, often arranged randomly and in a sparsely connected manner. The reservoir serves as a dynamic memory system, adept at capturing and processing temporal dependencies within the input data. ESN architecture (called RC henceforth) retains fixed parameters in both the input and reservoir layers while training the final linear layer, referred to as the "readout," utilizing least squared error with L2 norm regularization \cite{jaeger2001echo}.


Numerous research endeavors have delved into exploring the diverse applications of Reservoir Computing (RCs), with a particular emphasis on its utility in physical systems. A predominant focus within this realm has been the investigation of RCs for predicting the behavior of chaotic systems, exemplified by the examination of Lorenz systems in various studies \cite{chattopadhyay2020data, shougat2021reservoir, platt2022systematic}. Moreover, a substantial portion of the research landscape has been dedicated to the utilization of RCs in forecasting spatiotemporal patterns. Some of these investigations have expanded the scope of RC application, transitioning from low-dimensional to semi-high-dimensional forecasting tasks. For instance, studies have tackled challenges posed by datasets with spatially connected input variables, such as the Kuramoto-Sivashinsky dataset, where intricate relationships exist in a single dimension \cite{rafayelyan2020large, vlachas2019forecasting, vlachas2020backpropagation}. In a parallel line of inquiry, additional studies have ventured into the exploration of fully two-dimensional spatial datasets \cite{yao2022learning}, further enriching the understanding of RCs' efficacy in handling complex spatial relationships. The collective findings from these investigations contribute to a nuanced comprehension of the versatility and applicability of RCs, not only in predicting chaotic systems but also in addressing the challenges posed by spatiotemporal forecasting tasks across various dimensions. Despite the effectiveness of RCs, researchers have encountered a significant bottleneck hindering the comprehensive utilization of RCs in high-dimensional datasets, namely the exponential memory requirements essential for the effective training of RCs on such complex problems \cite{chattopadhyay2020data}. This challenge alongside higher computation requirements underscores the need for further advancements in addressing memory-related constraints to unlock the full potential of Reservoir Computing in handling high-dimensional datasets.

In the continuum of traditional deep learning progress, numerous iterations of RCs have emerged, capitalizing on the stratified architecture to either unveil multi-level temporal features \cite{gallicchio2017echo, gallicchio2017deep}, achieve a more sophisticated state-to-forecast representation through multiple readout layers \cite{pascanu2013construct, gallicchio2021deep} or concatenation of multiple input-reservoir-readout modules \cite{nichele2017deep}. Multiple studies \cite{gallicchio2017deep, gallicchio2017echo, goldmann2020deep} have demonstrated that increasing the number of reservoir layers with non-linear activations in a deep RC architecture, allows capturing dependencies across various time scales. The deep architecture is reported to enhance the internal representations formed during the neural encoding of input time-series data, thereby enriching the depth and complexity of the resultant model. 

Despite the proven effectiveness of sequential RC in capturing various temporal information, no study has investigated the benefits of this architecture in handling spatiotemporal patterns where a model requires learning multi-level spatial and temporal features. Motivated by the applications of sequential RC and recognizing the increasing demand for more nuanced spatiotemporal capabilities, we investigate this question along with the hardware efficiency of deep RC (hereafter called Sequential RC) models and compare them with the original RC models.

We experimented with the baseline and proposed sequential RC using two separate low- and high-dimensional data settings. We investigate a low-dimensional data called Lorenz63 which is a set of differential equations representing a simplified atmospheric convection model. Lorenz63 data falls under the chaotic dataset category where small errors in the current state can result in significant errors in near-term future states \cite{lorenz1963deterministic}. To assess performance on high-dimensional benchmarks, we utilized a comprehensive approach by employing two distinct Navier Stokes simulations. Specifically, these simulations focused on the dynamics of vorticity and the shallow water equations, both implemented within a two-dimensional configuration. The benchmark datasets allowed for a thorough exploration and analysis of physics-based behaviors, providing a robust and insightful evaluation of their performance in complex, multidimensional scenarios.

The subsequent sections of this paper are structured as outlined below: Section \ref{sec:method} provides a description of the methodology for both the baseline models and the proposed model. In Section \ref{sec:data}, we present the datasets used for low- and high-dimensional benchmarks. Section \ref{sec:results} showcases our experiments and analysis, elucidating the gains and defects in the model architecture across various settings. The paper concludes with a summary of our findings in Section \ref{sec:conclusion}.

\section{Methodology}\label{sec:method}
In this section, we present the approaches used in this paper. We present the two baselines of the recurrent neural networks (Elman-type and LSTM). Then introduce the proposed approach, reservoir computing, and describe the details of implementation.

\subsection{Recurrent Neural Networks}

Elman-type Recurrent Neural Networks (RNNs), commonly referred to as vanilla RNNs, constitute a distinct class of neural architectures highly adept at modeling sequential data. Elman-type RNNs, as introduced by \cite{elman1990finding}, stand out for their inclusion of a context unit in the hidden layer, enabling the network to learn and retain time-dependent information \cite{akbari2018short}. The core of the Elman-type RNN is its hidden layer, which contains recurrent connections. These connections allow the network to maintain an internal state or memory, enabling it to capture information from past time steps. The hidden layer is crucial for modeling short-term dependencies in sequential data. A distinguishing feature of the Elman-type RNN is the inclusion of a context or previous state. This layer has a one-on-one connection with the hidden layer, creating an inner loop in the network architecture. The context layer acts as a memory mechanism for time-dependent information, providing the network with the ability to learn and retain temporal patterns over longer sequences. However, it is important to note that traditional vanilla RNNs may encounter challenges when dealing with long-term dependencies, often due to vanishing or exploding gradient problems during training \cite{akbari2018short, gers2000learning}. This stems from the non-linear activation gradients squashing or magnifying the weights in the long loop of recurrent backpropagation. Mathematically speaking, given the input data ($x_t$), where $t \in [0, T]$, and the context state $h_{t-\Delta t}$, the model calculates the hidden state at time $t$ as follows:

\begin{equation}\label{eq:rnn}
    h_t = g(W_{x} x_t + W_{h} h_{t-\Delta t} + b_h)
\end{equation}
where \(W_{x}\) and \(W_{h}\) are weight matrices for the input and hidden states, respectively. \(b_h\) is the bias term and \(g\) is the non-linear activation function, commonly a hyperbolic tangent (tanh) or rectified linear unit (ReLU).

\subsection{Long Short-Term Memory}
Elman-type Recurrent Neural Networks (RNNs) are characterized by a well-documented challenge known as the gradient vanishing or exploding problem. The issue arises during the backpropagation through time (BPTT) process, specifically due to the involvement of multiple recursions in the network's architecture passing non-linear activation functions such as hyperbolic tangent and sigmoid. Consequently, certain gradients become either excessively small (vanishing) or disproportionately large (exploding), presenting a hindrance to effective training.
To address these challenges associated with Elman-type RNNs, advanced architectures such as Long Short-Term Memory (LSTM) networks and Gated Recurrent Unit (GRU) networks have been developed. These architectures incorporate specialized mechanisms, including gating units, to effectively manage the flow of information during training. By mitigating the vanishing and exploding gradient problems, these advanced models enhance the network's capability to capture and retain long-range dependencies within sequential data. LSTMs in particular, consist of three gating units: input gate ($i_t$), output gate ($o_t$), and late addition unit called forget gate ($f_t$) \cite{gers2000learning}. These gating units update each respective state similar to Elman-type RNN (Eq. \ref{eq:lstm_i}, \ref{eq:lstm_o}, \ref{eq:lstm_f}) independently and then input and forget gates share information to update the cell state ($c_t$). The cell state serves as a long-term memory unit that runs throughout the entire sequence. It allows LSTMs to capture and remember information over extended periods (Eq. \ref{eq:lstm_c}). Following the long-term memory, the hidden state, representing the short-term memory, is updated. It is influenced by both the current input and the previous hidden state. The hidden state ($h_t$) is used for making predictions and is updated based on the cell state and the output gate (Eq. \ref{eq:lstm_h}).

\begin{subequations}\label{eq:lstm}
    \begin{align}
        i_t &= \sigma(W_{ii} x_t + W_{hi} h_{t-\Delta t} + b_i) \label{eq:lstm_i}\\
        o_t &= \sigma(W_{io} x_t + W_{ho} h_{t-\Delta t} + b_o) \label{eq:lstm_o}\\
        f_t &= \sigma(W_{if} x_t + W_{hf} h_{t-\Delta t} + b_f) \label{eq:lstm_f}\\
        c_t &= f_t \odot c_{t-\Delta t} + i_t \odot \tanh(W_{ic} x_t + W_{hc} h_{t-\Delta t} + b_c) \label{eq:lstm_c}\\
        h_t &= o_t \odot \tanh(c_t) \label{eq:lstm_h}
    \end{align}
\end{subequations} 
where \(i_t\), \(o_t\), and \(f_t\) represent the input, output, and forget gates, respectively. \(\sigma\) is the sigmoid activation function and \(\odot\) represents element-wise multiplication. The weights for different components are named as $W_{[i/h][gate-first-letter]}$ where $i$ and $h$ represent the input layer and hidden layer, respectively. For instance, weights for the hidden layer of the forget gate are represented as $W_{if}$, or weights for the input layer of the cell state are named $W_{ic}$.

These equations collectively define the LSTM architecture, elaborating how the input, forget, and output gates, along with the cell state and hidden state, are updated and interact over timesteps in the network, preventing the underflow or overflow of gradient in the system sustaining a longer temporal comprehension of the data.






\subsection{Reservoir Computing}



Reservoir Computing (RC), initially proposed within the Echo-State Networks (ESNs) \cite{jaeger2001echo} and Liquid State Machines (LSMs) \cite{maass2002real} frameworks, has re-surged as a potent and efficient computational framework tailored for the modeling and forecasting of time-dependent sequences \cite{schrauwen2007overview, verstraeten2007experimental}. RCs, following the ESN model architecture, are comprised of two randomly initiated components namely input and reservoir layers that are fixed and not trained during the training phase. The input layer $W_{in} \in R^{D\times N}$ is in charge of the random transformation of input data to a latent representation which then is intake by the reservoir layer to derive dynamic temporal features via calculation of the states. The reservoir layer is a randomly squared matrix $W_{res} \in R^{N\times N}$ that serves as an adjacency matrix for the reservoir graph. The reservoir graph follows the Erdős–Rényi graph structure, where each graph edge has a certain probability of existence, \cite{han2022tighter} to ensure optimal connectivity. Experiments \cite{pathak2017using, pathak2018model} have demonstrated that, beyond the graph structure, reservoirs with a spectral radius approximately equal to unity preserve the echo state property and exhibit optimal performance \cite{pathak2017using}. Mathematically, given the data at time $t$ denoted as $x_t$, the previous state denoted as $r_{t - \Delta t}$, and the weights for the input and reservoir layers as $W_x$ and $W_r$, the state of the model can be obtained as:

\begin{equation}\label{eq:reservoir_classic_state}
    r_t = (1-\alpha) r_{t-\Delta t} + \alpha g\left(W_r r_{t-\Delta t} + W_x x_t \right)
\end{equation}

Here, $\alpha$ is the leak rate, governing the rate at which new information is incorporated into the system, and $g$ represents the non-linear function used as the activation function of the input layer.

\begin{figure*}[ht]
    \centering
    \includegraphics[width=0.6\textwidth]{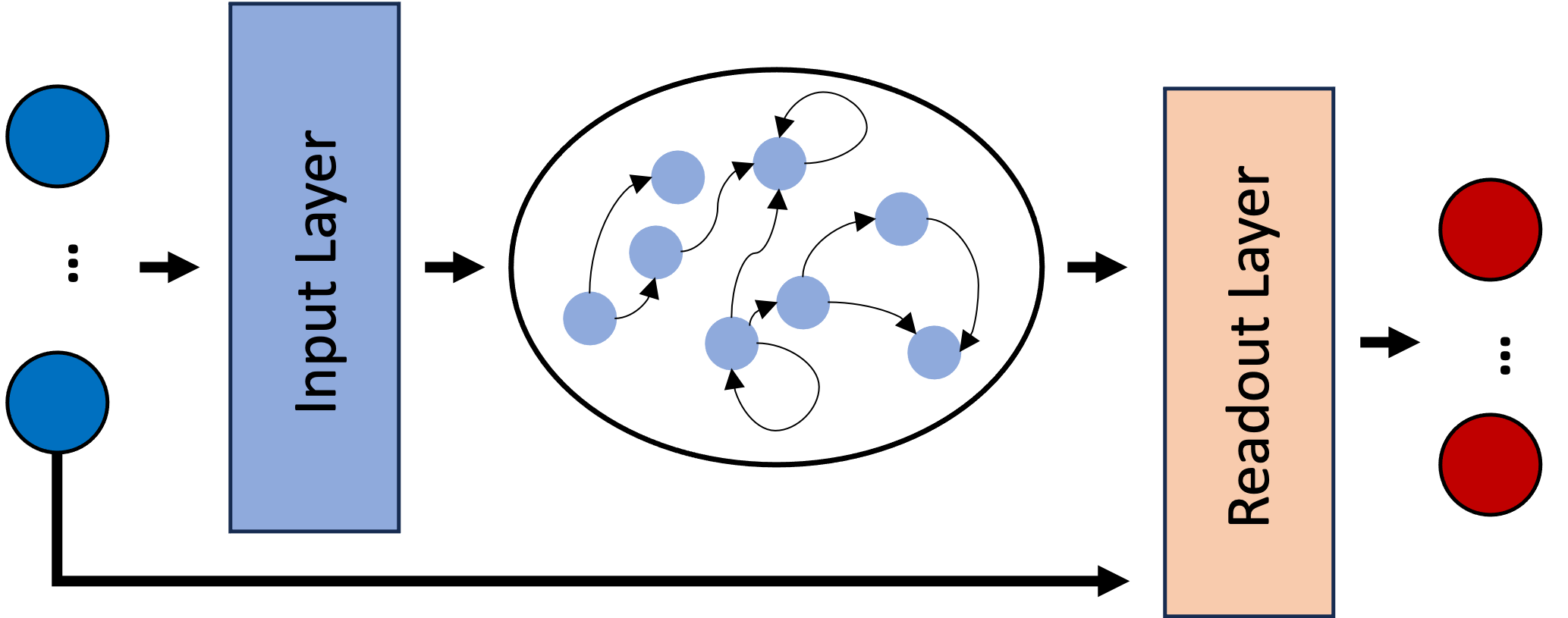}
    \caption{The schematic configuration of RC model architecture. The left-side blue nodes and right-side red nodes represent input and target variables, respectively.}
    \label{fig:rc_schema}
\end{figure*}

This architecture (Figure \ref{fig:rc_schema}) enables the model to capture intricate state patterns, contributing to effective and efficient forecasting. The model's states are calculated in an autoregressive manner and can subsequently be mapped to output data through an output layer known as the "readout" layer. The readout layer is essentially a multiple linear regression layer that maps the state vectors ($r_t$) to the output data ($y_t$). To improve generalization and prevent overfitting, the readout layer typically employs Tikhonov regularization, commonly referred to as Ridge regression. The objective of this regularization is to minimize the following cost function:

\begin{equation}\label{eq:reservoir_classic_readout1}
    \hat{W}_o = \arg \min_{W_o} \left\{ \lVert y - RW_o \rVert_2^2 + \beta \lVert W_o \rVert_2^2 \right\}
\end{equation}
where $W_o$ and $\hat{W}_o$ represent the readout parameters and their optimized versions, respectively. In this equation, $y$ denotes the target vector, $R$ represents the state vector \([1, x_t, r_t]\), and $\beta$ is the regularization parameter. The vectorized solution for Equation \eqref{eq:reservoir_classic_readout1} is given by:

\begin{equation}\label{eq:reservoir_classic_readout2}
    \hat{W_o} = (R^T R + \beta I)^{-1} R^T y
\end{equation}

Here, $I$ denotes the identity matrix. The integration of the processes described in Equations \eqref{eq:reservoir_classic_state} and \eqref{eq:reservoir_classic_readout2} constitutes the foundation for RC's ability to learn and predict dynamic temporal evolutions.

\subsection{Sequential Reservoir Computing}

\begin{figure*}[ht]
    \centering
    \includegraphics[width=\textwidth]{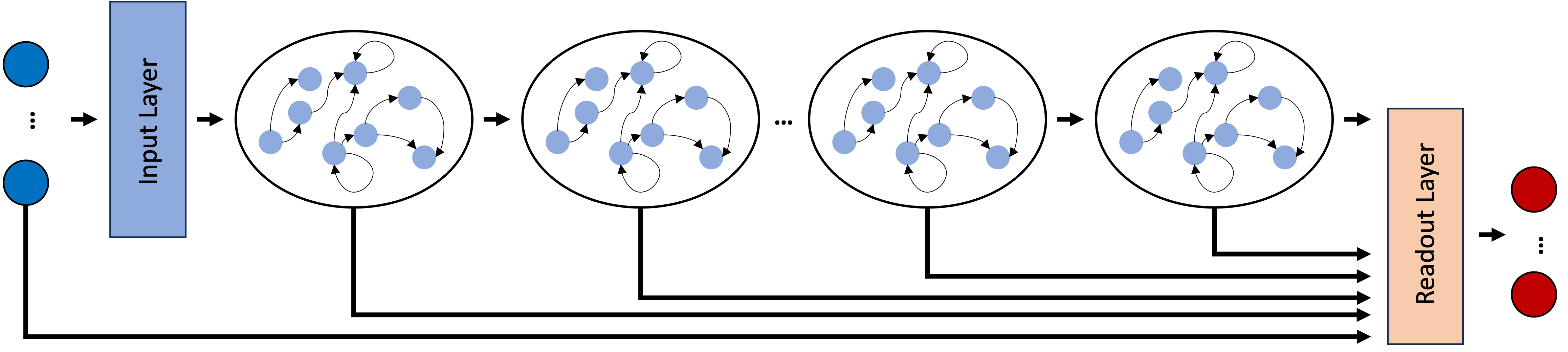}
    \caption{The schematic configuration of RC model architecture. The left-side blue nodes and right-side red nodes represent input and target variables, respectively. The outputs of each reservoir layer along with input data to the readout layer.}
    \label{fig:src_schema}
\end{figure*}

The proposed Sequential RC model (Figure \ref{fig:src_schema}) aims to enhance the capabilities of traditional reservoir computing by introducing multiple reservoir layers connected in sequence. In this model, each reservoir layer processes the subsequent layer's information and the output of each reservoir is concatenated before being fed into a readout layer for final prediction (Figure \ref{fig:src_schema}). The sequential arrangement of reservoir layers allows the model to capture complex temporal dependencies and improve overall performance with a significantly smaller number of reservoir parameters that can become the bottleneck of scaling to high-dimensional data. In mathematical form, Sequential RC consists of an input layer with parameters denoted as \(W_{x}\), multiple reservoir layers, denoted as \(W_{r}^1, W_{r}^2, ..., W_{r}^n\), are sequentially connected. Each reservoir layer \(W_{r}^i\) has its own internal dynamics and weights. The dynamics of reservoir \(W_{r}^i\) can be described by the following equations:

\begin{equation}
    r_t^i = \begin{cases} 
    (1-\alpha) r_{t-\Delta t}^i + \alpha g\left(W_r^i r_{t-\Delta t}^i + W_x x_t \right), & \text{for } i = 1 \\
    (1-\alpha) r_{t-\Delta t}^i + \alpha g\left(W_r^i r_{t-\Delta t}^i + W_r^{i-1} r_{t}^{i-1} \right), & \text{for } i \neq 1
    \end{cases}    
\end{equation}
where \(i\) represents the reservoir layer order \( i = 1, \ldots, n \) with \(n\) representing the total number of reservoir layers.

Similar to the RC notation, the output of each reservoir layer, denoted as \(\mathbf{r}_t^i\), is concatenated for all reservoir layers to form the input for the readout layer. The concatenated output is represented as \(\mathbf{r}^{\text{concat}}_t\):

\begin{subequations}
    \begin{align}
        R_t = \mathbf{r}^{\text{concat}}_t = [1, x_t, r_t^1, r_t^2, ..., r_t^n]\\
        \hat{W_o} = (R^T R + \beta I)^{-1} R^T y \label{eq:SRC_readout}
    \end{align}
\end{subequations}

The readout layer parameters for Sequential RC are calculated in a similar way to RC as shown in Eq. \ref{eq:SRC_readout}.

\section{Datasets}\label{sec:data}

In this section, we will go over the datasets employed in our experimental investigations, providing a detailed account of their respective attributes, and expound upon the specific hyperparameters instantiated for these experiments.

\subsection{Lorenz63}
The Lorenz63 system is a set of ordinary differential equations that serves as a classic example of a chaotic dynamical system. It was introduced by Edward N. Lorenz in 1963 \cite{lorenz1963deterministic} as a simplified atmospheric convection model. The system comprises three coupled nonlinear differential equations, representing the evolution of three state variables. The Lorenz63 equations are given by:
\begin{subequations}
    \begin{align}
        \frac{dx}{dt} &= \sigma(y - x) \\
        \frac{dy}{dt} &= x(\rho - z) - y \\
        \frac{dz}{dt} &= xy - \beta z
    \end{align}
\end{subequations}
where $x$ $y$, and $z$ are the state variables, and $\sigma$, $\rho$, and $\beta$ are parameters governing the system's behavior. The Lorenz63 system is known for its sensitivity to initial conditions, leading to chaotic trajectories in its phase space. Due to its simplicity yet rich dynamics, the Lorenz63 system has become a standard testbed for studying chaos, attractors, and the behavior of dynamical systems in diverse scientific disciplines, including atmospheric science, physics, and mathematics. In our experiments, we leverage the Lorenz63 dataset to explore and analyze the complex dynamics inherent in chaotic systems as our low-dimensional benchmark dataset. We initialized the dynamics using values 17.67715816276679, 12.931379185960404, 43.91404334248268 respectively for $\sigma$, $\rho$, and $\beta$.


\subsection{Vorticity}

We examine the 2D Navier-Stokes equation characterizing the dynamics of a viscous behavior of an incompressible fluid expressed in vorticity form on a torus-shaped region. The governing equations are given by:

\begin{subequations}
    \begin{align}
        \partial_t w(x, t) + u(x, t) \cdot \nabla w(x, t) &= \nu \Delta w(x, t) + f(x), \quad x \in (0, 1)^2, \quad t \in (0, T], \label{eq:ns1}\\
        \nabla \cdot u(x, t) &= 0, \quad x \in (0, 1)^2, \quad t \in [0, T], \label{eq:ns2}\\ 
        w(x, 0) &= w_0(x), \quad x \in (0, 1)^2 \label{eq:ns3}
    \end{align}
\end{subequations}

In the equations \ref{eq:ns1}-\ref{eq:ns3}, \(w(x, t)\), \(u(x, t)\), \(\nu\) and \(f(x)\) represents vorticity's evolution over time, fluid's velocity, viscosity, and an external force, respectively. The velocity is derived from the vorticity, and the whole system is constrained by the incompressibility condition (Eq. \ref{eq:ns2}). Initial condition \(W_{0}(x)\) is generated according to \(w_{0} ~ \mathcal{N}(0, 7^{3/2}(-\Delta+49I)^{-2.5}\) with periodic boundary conditions \cite{li2020fourier}. We used a fixed forcing \(f(x) = (sin(2\pi(x_{1} + x_{2})) + cos(2\pi(x_{1} + x_{2}))) / 10 \). For more information on the solver specifications and assumptions, please refer to \cite{li2020fourier}. In our experiments, we used a fixed Reynolds Number of 500 influencing the viscosity (\(\nu\)) in Eq. \ref{eq:ns1}. The resolution of our simulation grid is fixed at 64 × 64 for consistency in training and testing, as some existing methods are sensitive to resolution changes.

\subsection{Shallow Water Equations}

The shallow-water equations (SWEs) form a fundamental set of equations in oceanography, hydrology, and related fields. They describe the dynamics of free-surface flow in a thin layer, where the water depth is much smaller than the horizontal length scales. Solving the shallow water equations in two dimensions, a canonical system that effectively captures the dynamics of fluid motion in scenarios where the fluid depth is substantially smaller than its horizontal extent. The governing equations for this model encapsulate both the linearized momentum equations and the non-linearly solved continuity equation. The linearization of the momentum equations allows for a tractable representation of fluid motion, facilitating numerical simulations of wave propagation and interactions, while the non-linear solution to the continuity equation ensures the preservation of mass within the system. The model's configuration initiates with a prominent Gaussian bump, serving as an initial disturbance in the water surface, leading to the generation of waves that propagate away from the disturbance. Subsequently, the waves interact with the model boundaries, adhering to a no-flow condition.

The model incorporates linear momentum equations alongside the solution of the continuity equation in its nonlinear form. The flexibility of the script allows for the selective activation or deactivation of various terms. In its most comprehensive configuration, the model is governed by the following set of equations:


\begin{subequations}
    \begin{alignat}{2}
        \frac{\partial u}{\partial t} - fv &= -g\frac{\partial \eta}{\partial x} + \frac{\tau_x}{\rho_0 H} - \kappa u \\
        \frac{\partial v}{\partial t} + fu &= -g\frac{\partial \eta}{\partial y} + \frac{\tau_y}{\rho_0 H} - \kappa v \\
        \frac{\partial \eta}{\partial t} + \frac{\partial (\eta + H)u}{\partial x} + \frac{\partial (\eta + H)v}{\partial y} &= \sigma - w
    \end{alignat}
\end{subequations}
where \(f = f_0 + \beta y\) represents the full latitude-varying Coriolis parameter. The momentum equations utilize an ordinary forward-in-time centered-in-space scheme. However, handling the non-trivial Coriolis terms involves a two-step process: predicting values for the horizontal velocity components in the x- and y-directions (\(u\) and \(v\)) and subsequently correcting them to incorporate the Coriolis effects. The continuity equation employs a forward difference for the time derivative and an upwind scheme for the nonlinear terms.

The stability analysis of the model is conducted under the Courant-Friedrichs-Lewy (CFL) condition, ensuring that the time step (\(dt\)) is constrained by \(dt \leq \min(dx, dy) / \sqrt{gH}\). Additionally, stability requires that \(\alpha \ll 1\) when Coriolis effects are considered, where \(dx\) and \(dy\) represent the grid spacing in the x- and y-directions, \(g\) is the acceleration due to gravity, and \(H\) is the resting depth of the fluid. \(\eta\), \(g\), \(\rho_{0}\), \(\kappa\), \(\sigma\), \(w\), \(\tau_x\) and \(\tau_y\) represent the surface height, gravity, fluid density, velocity coefficient, additional source/sink term, vertical velocity, external forcing terms in the x- and y-directions, respectively.

\section{Results and Discussion}\label{sec:results}

\subsection{Metrics}

\subsubsection{Valid Prediction Time}
We employ the valid prediction time (VPT) \cite{vlachas2019forecasting, vlachas2020backpropagation, platt2022systematic} to track the forecast extent directly in low-dimensional time-series data (i.e. Lorenz63). VPT is defined as the time instance \(t\) when the deviation between simulated predictions and the ground truth surpasses a predefined threshold based on the root mean square error (RMSE):

\begin{subequations}
    \begin{align}
        \text{RMSE}(t) \geq \varepsilon\\
        \text{RMSE}(t) = \sqrt{\frac{1}{D} \sum_{i=1}^{D} \left(\frac{\hat{y}_i(t) - y_i(t)}{\sigma_i}\right)^2}
    \end{align}
\end{subequations}

The \(i\)-th component at time \(t\) for predictions (\(\hat{y}_i(t)\)) and the corresponding ground truth (\(y_i(t)\)) are denoted. The chosen threshold (\(\varepsilon\)) for this study is set to 0.3. Within the RMSE metric, \(\sigma_i\) signifies the \(i\)-th component of the standard deviation in the true data, serving a normalization purpose. The symbol \(D\) represents the dimensionality of the problem under consideration.

\subsubsection{Return Map}\label{sec:return_map}

The Return Map is a concept used in the study of dynamical systems introduced by Lorenz \cite{lorenz1963deterministic} to analyze the stability of a simulation, particularly by observing the dynamics of Lorenz63. Lorenz observed that a local maxima in variable \(z(t)\) occurs approximately when a circuit (a full circle in Lorenz63's butterfly shape) completes. Analyzing the successive local maxima in \(z(t)\) provides insights into the general alignment of predictions with the groundtruth. Following Lorenz's recipe, we calculated the maximum values of \(z(t)\) in its immediate neighborhood and then plotted the \(z_{i}(t)\) versus \(z_{i+1}(t)\) to generate the return map curve.

\subsubsection{Peak Signal-to-Noise Ratio}
Peak Signal-to-Noise Ratio (PSNR) is a widely utilized quality assessment metric in image and video processing, providing a quantitative measure of the fidelity between an original and a distorted signal. Mathematically, it is expressed as the logarithmic ratio of the maximum possible intensity value squared to the mean squared error between the original and distorted signals. Specifically, for image comparison, the PSNR is given by \(PSNR = 10 \cdot \log_{10}\left(\frac{{\text{{max}}^2}}{{MSE}}\right)\), where \(\text{{max}}\) represents the maximum possible pixel intensity value, and MSE denotes the Mean Squared Error. A higher PSNR value indicates a smaller distortion and, consequently, a higher quality of the reconstructed signal. Despite its widespread use, it is essential to acknowledge that PSNR may not perfectly align with perceived image quality due to its reliance on mean squared errors and insensitivity to certain perceptual nuances. Nonetheless, it remains a valuable quantitative tool for assessing the fidelity of reconstructed signals in various image and video processing applications.

\subsubsection{Structural Similarity Index}
Structural Similarity Index (SSIM) is a comprehensive metric employed in image processing to quantify the similarity between an original (\(X\)) and a distorted (\(Y\)) image. Unlike traditional metrics such as RMSE or PSNR, SSIM considers not only luminance differences but also incorporates information about structural patterns, textures, and contrasts perceptible to the human visual system. The SSIM index (\(SSIM(X,Y)\)) is calculated by evaluating three components: luminance (\(l(X,Y)\)), contrast (\(c(X,Y)\)), and structure (\(s(X,Y)\)). These components are combined using the following formula:

\begin{equation}
    SSIM(X,Y) = \frac{{[l(X,Y) \cdot c(X,Y) \cdot s(X,Y)]^\alpha}}{{[l(X,Y)^{\prime\prime} \cdot c(X,Y)^{\prime\prime} \cdot s(X,Y)^{\prime\prime} ]^\beta}}.
\end{equation}

Here, \(\alpha\), \(\beta\), and \(\gamma\) are constants to stabilize the division, and \(l(X,Y)\), \(c(X,Y)\), and \(s(X,Y)\) are local mean, standard deviation, and cross-covariance functions, respectively. The SSIM index yields a value ranging from -1 to 1, where 1 indicates perfect similarity. This metric is particularly advantageous in scenarios where human perception plays a crucial role, as it aligns more closely with perceived image quality. It has found widespread use in the evaluation of image and video compression algorithms, image restoration techniques, and other applications where the preservation of structural information is essential for maintaining visual fidelity.

\subsection{Low-dimensional Results}

In this section of our study, we focus on low-dimensional data, specifically the Lorenz63 system, as the basis for our experiments. Our objective is to assess the performance of baseline models and draw comparisons with RC and Sequential RC models. To delve into the influence of training data on forecasting capabilities and to comprehend potential advantages offered by less parameterized models, such as RC and Sequential RC, we conduct experiments with three distinct training data lengths. These scenarios involve training our models with 2,000, 5,000, and 10,000 samples, enabling us to examine the impact of varying training dataset sizes. Additionally, we extend our exploration to assess the efficiency of these models by scrutinizing their number of trainable and fixed parameters, Floating Point Operations Per Second (FLOPS), memory usage and training time, providing valuable insights into their computational efficacy. Through this multifaceted analysis, we aim to contribute to a nuanced understanding of the dynamics between training data, model complexity, and forecasting efficiency in low-dimensional systems.

\begin{figure*}[htbp]
    \centering
    \includegraphics[width=\textwidth]{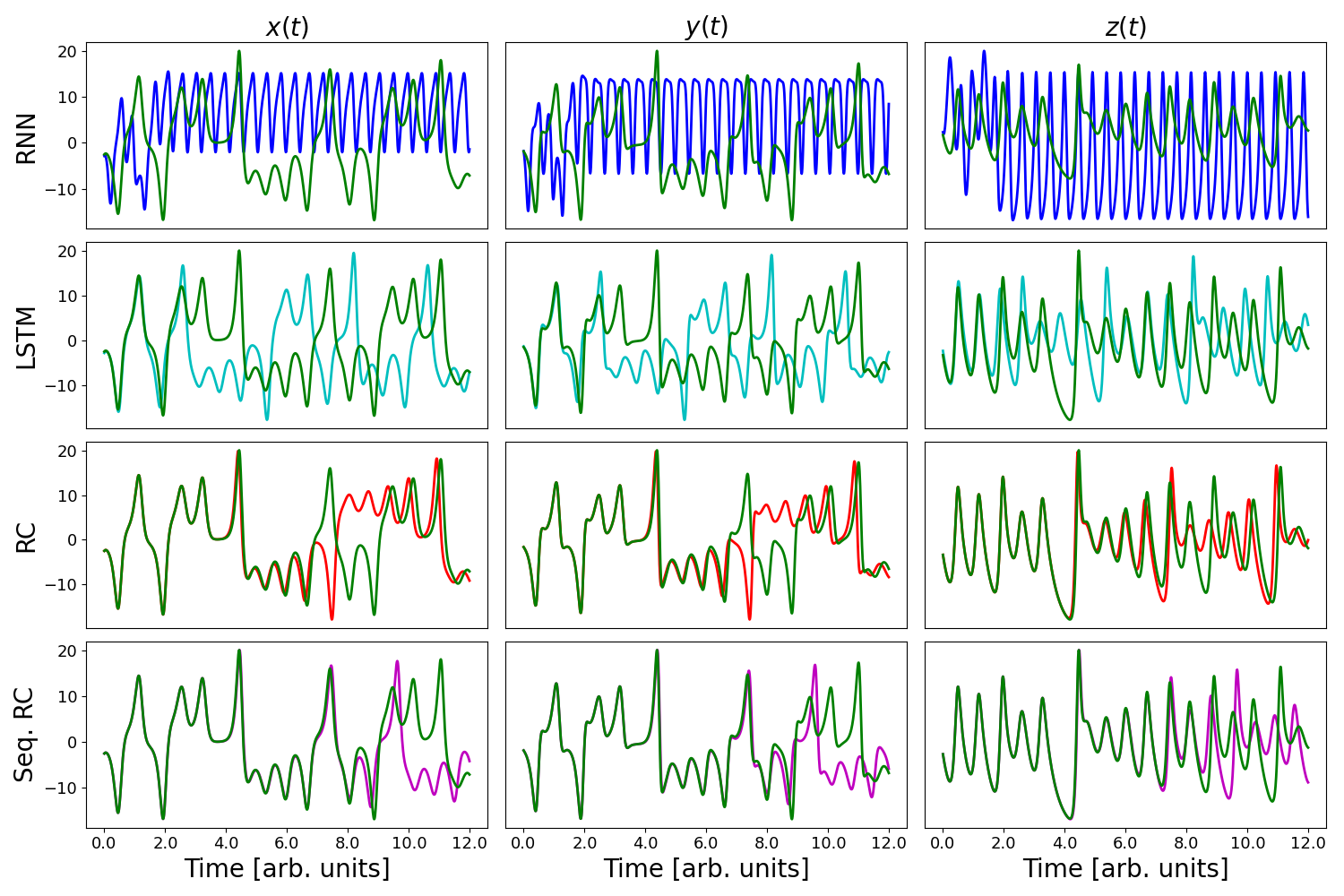}
    \caption{Forecasts based on 2,000 training samples. The y-axis in each subplot represents the respective variable's value and the x-axis represents the lead-time of forecast in arbitrary time units.}
    \label{fig:res2000}
\end{figure*}

In the first experiment, we trained the many-to-one architecture RNN and LSTM models with a fully-connected input layer, one RNN/LSTM layer, and the output layer to narrow the architectural differences between the baselines and the proposed RC and Sequential RC. The choice of many-to-one architecture is to prepare the relevant context for the RNN/LSTM forecast, similar to the RC/Sequential RC context role in the forecast. We performed a hyper-parameter search of the length of the historical window effect on the model performance (See Appendix \ref{sec:appendix_a}). The hyper-parameter search indicates that model performance increases by including a longer historical window in exchange for a higher computational burden. We chose a historical window of 49 based on our observations in Figure \ref{fig:hyperparameter_lookback} in \ref{sec:appendix_a}, and optimized both models using RMSProp \cite{tieleman2012lecture} of 1e-4 learning rate. Our results for RNN, LSTM, and RC are based on a recurrent layer/reservoir of size 256, and for Sequential RC we used 8 reservoir layers of size 32. Figure \ref{fig:res2000} demonstrates the model performances based on 2,000 training samples (for the 5,000 and 10,000 training samples, please refer to Appendix \ref{tab:model_train_time_lorenz}). The results show limited performances for both RNN and LSTM models for autoregressive forecasting whereas RC and Sequential RC show significant performance in terms of accurate forecasting time. It is noteworthy that Sequential RC is maintaining forecasting performance for a longer time despite the more limiting architecture than RC. In particular, Sequential RC has the same number of trainable parameters, and less than 14 percent of fixed parameters (Table \ref{tab:model_train_time_lorenz}).

In figure \ref{fig:res2000}, rows from top to bottom represent RNN, LSTM, RC, and Sequential RC models, respectively, and from left to right, columns represent \(x(t)\), \(y(t)\) and \(z(t)\) variables. The RNN model (in blue lines) shows weak forecasting skills and deviates from groundtruth (green line) early on in the forecast horizon and afterward starts oscillating. LSTM (cyan line in the second row) stays consistent with groundtruth for a longer period compared to RNN but starts deviating from the groundtruth from 1.73 VPT. RC and Sequential RC can both maintain forecast skills for a longer time than backpropagation-based models under the small training sample size. Sequential RC is able to forecast up to 8.65 VPT which is more than RC's 7.19 VPT. With the increase of the number of training samples from 2,000 to 5,000 and 10,000 similar performances are demonstrated except for LSTM where an increase of training samples from 2,000 to 5,000 approximately doubles the VPT. The suboptimal performance of backpropagation-based approaches is linked to the explosion of gradients while learning chaotic behavior \cite{mikhaeil2022difficulty}. The paper proposes strategies for addressing challenges in training recurrent models on chaotic datasets; however, this topic is beyond the scope of our study. Furthermore, the performance results for both RC and Sequential RC are derived from configurations with a spectral radius of 1.1, a leak rate of 0.7, and no sparsity for the 2,000 training samples (see Appendix \ref{sec:appendix_b} for model specifications). The performance of the four models concerning VPT, across various sample sizes, is illustrated in Table \ref{tab:VPT_lorenz}. Notably, in all training sample sizes, Sequential RC consistently matches or surpasses the performance of RC. 

\begin{table}[htbp]
    \centering
    \begin{tabular}{lcccc}
        \toprule
        Training Samples & RNN & LSTM & RC & Seq. RC \\
        \midrule
        2,000 & 0.10 & 1.73 & 7.19 & 8.65 \\
        5,000 & 0.24 & 3.11 & 8.58 & 8.57 \\
        10,000 & 0.17 & 3.17 & 7.96 & 8.67 \\
        \bottomrule
    \end{tabular}
    \caption{Temporal Prediction Validity Durations (VPT) in arbitrary time units for each model across different dataset sizes in the Lorenz63 experiment.}
    \label{tab:VPT_lorenz}
\end{table}

\begin{figure*}[ht]
    \centering
    \includegraphics[width=\textwidth]{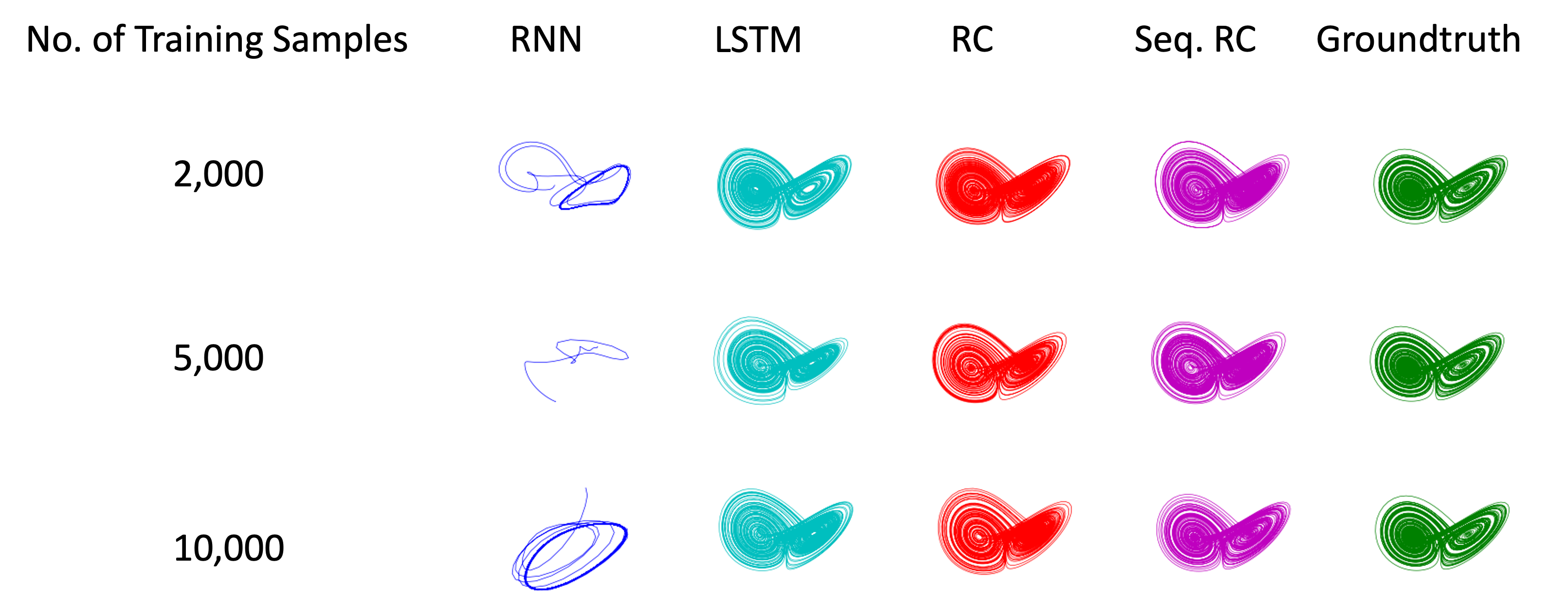}
    \caption{Comparison of Lorenz attractors for RNN (left column in blue), LSTM (second column from left in cyan), RC (third column from left in red), and Sequential RC (fourth column from left in magenta) and groundtruth (far right column in green). The Lorenz63 attractors illustrate the dynamic behavior of the systems, with differences in trajectory patterns and stability. The RNN is not able to follow the trajectory with any number of training samples. LSTM can't learn the trajectory but is able to pick the trajectory with 5,000 and 10,000 samples. RC and Sequential RC both are capable of forecasting trajectory in all training sample sizes.}
    \label{fig:attractors}
\end{figure*}

Additionally, Figure \ref{fig:attractors} indicates that, apart from RNN, all models exhibit the ability to learn chaotic dynamics across different training sample sizes. 

We plotted the return maps described in section \ref{sec:return_map} to observe the stability of the long-range dynamics. Results show no convergence between the groundtruth and RNN for any training sample size. LSTM shows better convergence but is not completely aligned with the groundtruth return map. It is noteworthy that LSTM return map convergence improves significantly going from 2,000 training samples to 5,000. RC and sequential RC demonstrate close alignment to groundtruth return map in all training sample sizes.

\begin{figure*}[htbp]
    \centering
    \includegraphics[width=\textwidth]{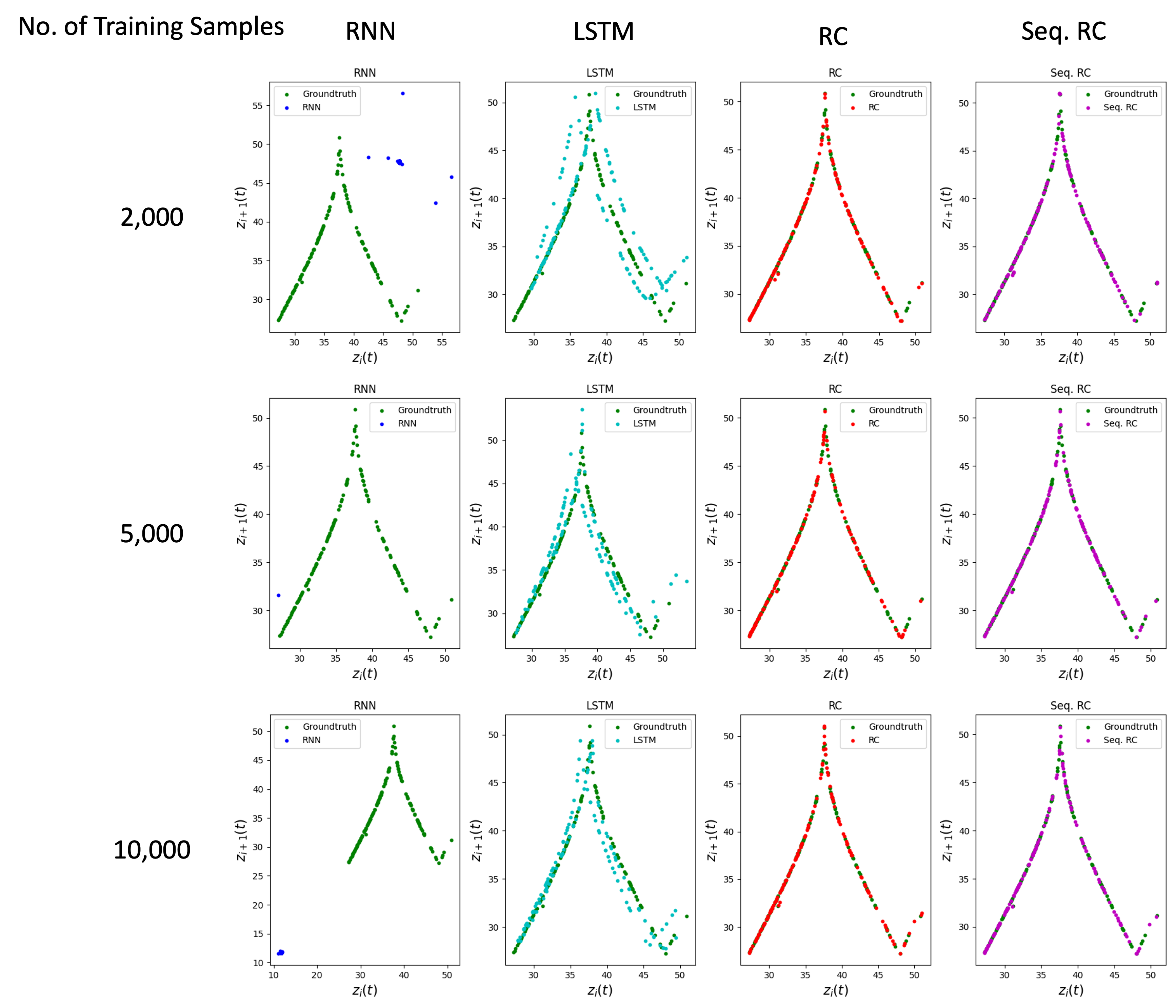}
    \caption{Return map for the \(z\)-dimension of the Lorenz63 system. The plot illustrates that RC and sequential RC models closely match the ground truth, showcasing their robust performance in all training sample sizes. In contrast, the RNN exhibits divergence, while the LSTM aligns closely but doesn't fully capture the dynamics. The comparison underscores the effectiveness of RC and sequential RC in maintaining stability while capturing the Lorenz63 dynamics.}
    \label{fig:return_map}
\end{figure*}

The computational and memory demands of the models employed in this study play a crucial role in understanding their efficiency and scalability. All the models discussed herein are trained on a single Graphics Processing Unit (GPU) to ensure maximum fairness in the evaluation process despite the known limited GPU-accelerated operations in RC and sequential RC. The reported memory metrics specifically refer to GPU memory usage. In terms of memory demand, GPU memory utilization is a critical factor. The memory demand is influenced by the model architecture, the size of the dataset, and the complexity of the tasks. GPU memory is crucial for storing intermediate results, gradients, and parameters during the training process. Understanding and monitoring GPU memory usage is vital to ensure that the chosen GPU has sufficient capacity to handle the computational load.

In the interest of fairness, all experiments reported in this study are conducted on an NVIDIA GeForce RTX 3090 and 4 Intel(R) Xeon(R) E5-1660 v4 @ 3.20GHz cpus. The reported memory metrics, in table \ref{tab:model_train_time_lorenz}, exclusively pertain to GPU memory usage, providing insights into the resource requirements of the models and aiding in the assessment of their practical feasibility in real-world applications. The results presented in Table \ref{tab:model_train_time_lorenz} demonstrate the model parameters, inference computation and memory burden, and training time for each of the models. The outcomes reveal that in comparison to the RNN, LSTM, and RC, the sequential RC exhibits significantly fewer operations, with a reduction factor of 724.5, 2885, and 7.3, respectively. In a parallel pattern, the sequential RC demonstrates memory efficiency with a reduction factor of 111.8, 471.7, and 1.5 in comparison to RNN, LSTM, and RC, respectively. When it comes to training time, sequential RC falls short of surpassing RC's training time due to the sequential nature of calculations and the associated latency. Nevertheless, it remains comparable and advantageous as the memory consumption of RCs experiences an exponential increase, while sequential RCs exhibit notably lower memory requirements.

        

\begin{table*}[htbp]
    \centering
    \begin{tabular}{lccccc}
        \toprule
        Metric & {No. Training Samples} & {RNN} & {LSTM} & {RC} & {Seq. RC} \\
        \midrule
        Trainable Parameters & {-} & {133,379} & {528,131} & {780} & {780}\\
        Fixed Parameters & {-} & {0} & {0} & {66,304} & {8,288}\\
        FLOPS (MFLOPS) & {-} & {6.5927} & {26.2535} & {0.0673} & {0.0091}\\
        Memory Consumption (MB) & {-} & {1.0176} & {4.2930} & {0.0138} & {0.0069}\\
        \midrule
        Training Time (s) & {2,000} & {508.0} & {576.0} & {0.4} & {1.4} \\
        & {5,000} & {1290.0} & {1452.0} & {1.1} & {3.9} \\
        & {10,000} & {2588.0} & {2932.0} & {2.2} & {7.7}\\
        \bottomrule
    \end{tabular}
    \caption{The table represents model statistics in terms of numbers of trainable and fixed parameters and MegaFLOPS (MFLOPS) and GPU memory consumption for a single forward run. In addition to the model statistics, training time for each model across different dataset sizes in the Lorenz63 experiment is calculated in seconds using an NVIDIA GeForce RTX 3090 and 4 Intel(R) Xeon(R) E5-1660 v4 @ 3.20GHz cpus.}
    \label{tab:model_train_time_lorenz}
\end{table*}

\subsection{High-dimensional Results}

In this section of our investigation, our attention shifts toward high-dimensional datasets, specifically the Vorticity and SWE datasets introduced in Section \ref{sec:data}. Drawing parallels with our exploration of low-dimensional data, our objective is to analyze the performance of baseline models and make comparisons with RC and sequential RC models within this high-dimensional context. As we delve into the higher-dimensional non-chaotic datasets, we aim to unravel the potential advantages offered by less parameterized models, such as RC and sequential RC. Our experiments encompass varied training data lengths, including 2,000, and 5,000 samples, allowing us to investigate the impact of training dataset sizes on model performance. Additionally, similar to the low-dimensional subsection, we extend our investigation to assess the efficiency of these models by evaluating their number of trainable and fixed parameters, FLOPS, memory usage, and training time. RNN, LSTM, and RC are designed to have a similar architecture to the Lorenz63 models, each featuring one hidden layer with a size of 512. For sequential RC, we employed 8 reservoir layers, each with a size of 64. Both RNN and LSTM were trained using RMSProp with a learning rate of $10^{-4}$. For RC and sequential RC, we used a spectral radius of 1.1, a leak rate of 0.7, and no sparsity.

\begin{figure*}[htbp]
    \centering
    \includegraphics[width=1.1\textwidth]{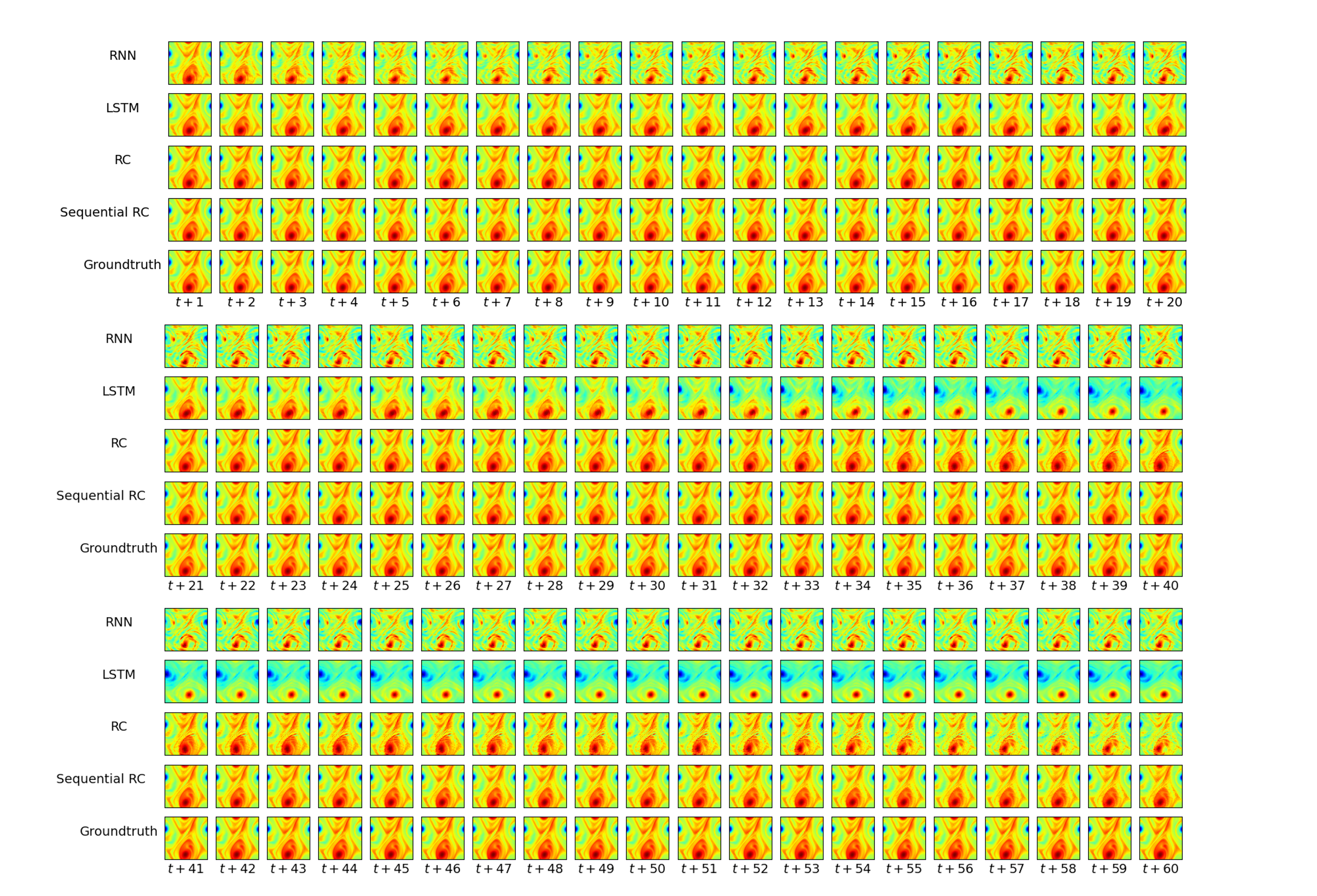}
    \caption{The figure provides a visual representation of forecasts for each model, arranged in rows (RNN, LSTM, RC, sequential RC, and groundtruth) and columns depicting forecast time steps ($t+1$ to $t+60$). Progressing from top to bottom within each block, the subplots showcase incremental forecast time intervals. Despite the seemingly slow progression due to small \(\Delta t\), crucial patterns emerge. RNN diverges from the initial steps, LSTM diverges around $t+30$, RC starts losing fine structure by $t+50$, and sequential RC maintains forecasts even at later time steps. The illustration captures the varying capabilities of each model in forecasting the Vorticity data.}
    \label{fig:vorticity_vis}
\end{figure*}

We first investigate the visual results of the baseline models and proposed sequential RC for the Vorticity dataset (Figure \ref{fig:vorticity_vis}). In Figure \ref{fig:vorticity_vis}, we present a visual analysis of model forecasts for the Vorticity dataset with a 2,000 training sample size. Each row corresponds to a specific model (RNN, LSTM, RC, sequential RC, and groundtruth), while columns represent consecutive forecast time steps ranging from $t+1$ to $t+60$. The subplots progress slowly with small \( \Delta t\) intervals, providing a nuanced perspective on the models' forecasting capabilities. Observations from the visual comparison show that the RNN exhibits divergence from the initial forecast steps, signaling its limited ability to capture the system's dynamics effectively. In the case of LSTM, divergence becomes apparent around $t+30$, highlighting challenges in maintaining accurate forecasts over time. The RC model, while initially effective, starts to lose fine structure in the forecasts by $t+50$. The sequential RC model stands out in its ability to maintain forecasts even at later time steps. The progression of subplots illustrates its stability and effectiveness in capturing the intricate patterns of the Vorticity dataset. 

\begin{figure}[htbp]
    \centering
    \includegraphics[width=0.8\columnwidth]{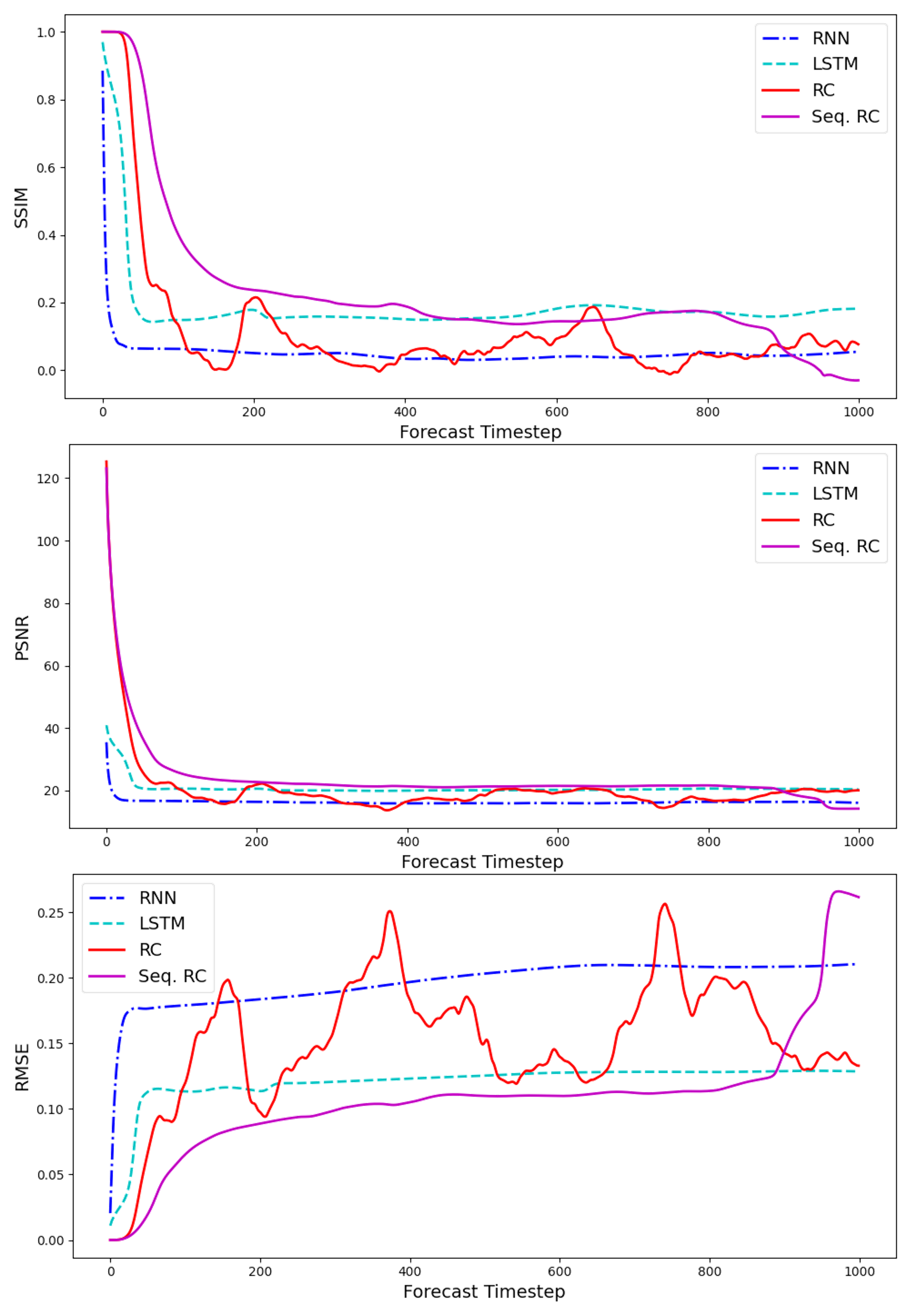} 
    \caption{Quantitative evaluation of forecasting models: SSIM, PSNR, and RMSE metrics plotted against forecast lead-time. The Y-axis represents metric values, and the X-axis denotes forecast lead-time from 1. Notably, RNN exhibits early deviation from ground truth, indicating limited forecasting proficiency. While LSTM degrades more slowly, it faces declining accuracy in early lead-times. RC and sequential RC initially perform comparably, but after around 50 timesteps, RC experiences a decline in SSIM and PSNR with a simultaneous rise in RMSE. In contrast, sequential RC maintains acceptable forecasting skills until the 94th timestep, highlighting its resilience and sustained performance over extended lead-times.}
    \label{fig:vorticity_graph}
\end{figure}

To complement our visual analysis, we conducted a quantitative assessment using SSIM, PSNR, and RMSE, as depicted in Figure \ref{fig:vorticity_graph}. In these plots, the Y-axis represents the metric value, while the X-axis signifies the forecast lead-time, starting from 1. As illustrated in Figure \ref{fig:vorticity_graph}, the RNN consistently deviates from the groundtruth across all metrics early in the forecast horizon, showcasing its limited forecasting skills. The LSTM model exhibits a slower degradation from reality across all metrics, yet it still experiences a decline in forecasting performance early in the lead-time. Both RC and sequential RC initially perform closely, demonstrating comparable forecasting skills in the early stages of the forecast lead-time. However, after approximately 50 timesteps, the RC model experiences a notable decrease in SSIM and PSNR, coupled with a sudden increase in RMSE, indicating a decline in forecasting skills. In contrast, sequential RC maintains acceptable forecasting skills until the 94th timestep, showcasing its robustness and sustained performance over a more extended lead-time. This quantitative evaluation augments our visual findings, providing a perspective on the forecasting capabilities of each model across varying lead-times.

\begin{figure*}[!htbp]
    \centering
    \includegraphics[width=1.1\textwidth]{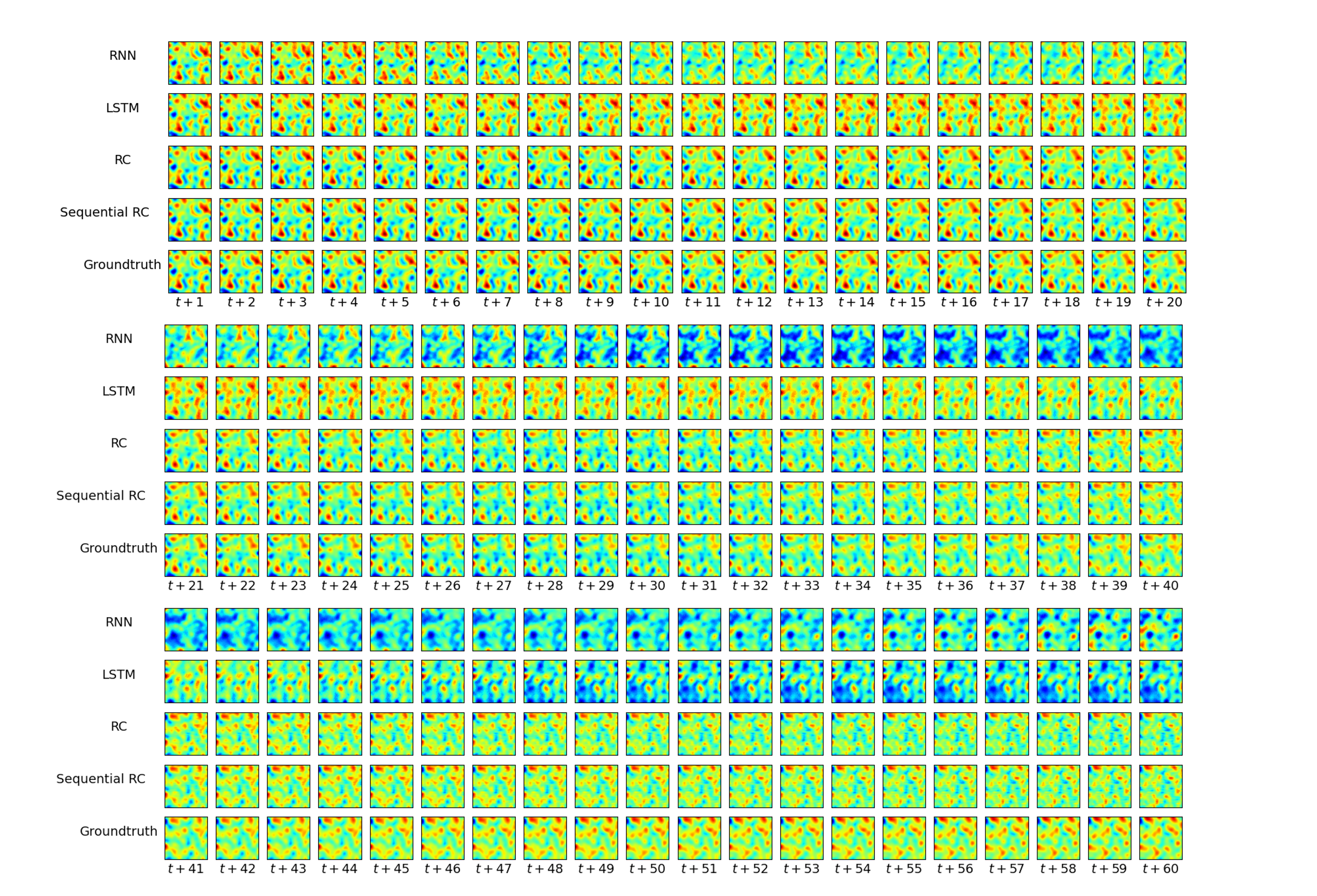}
    \caption{Visual analysis of model forecasts for the SWE dataset with a training sample size of 2,000. Each row corresponds to a specific model, and columns represent consecutive forecast time steps. The subplots exhibit SWE dynamics with confined boundaries, showcasing patterns analogous to the Vorticity dataset. Notably, the RNN displays early divergence from the ground truth, while the LSTM model exhibits sustained performance before divergence around \(t+20\). The RC and sequential RC models preserve dynamics for a longer range of forecasts, deteriorating around \(t+48\).}
    \label{fig:swe_vis}
\end{figure*}

Similarly, for the SWE dataset, we examine visual results for both baseline models and the proposed sequential RC, as depicted in Figures \ref{fig:swe_vis}. In this visual analysis, we explore the model forecasts for the SWE dataset with a specific focus on a training sample size of 2,000. The figure depicts rows corresponding to distinct models, including RNN, LSTM, RC, sequential RC, and groundtruth. The columns represent consecutive forecast time steps categorized into three blocks: \(t+1\) to \(t+20\) in the top block, \(t+21\) to \(t+40\) in the middle block, and \(t+41\) to \(t+60\) in the bottom block. The subplots within the figure showcase the dynamics of SWE, maintaining confined boundaries as detailed in Section \ref{sec:data}. Upon visual inspection, discernible patterns similar to those observed in the Vorticity dataset emerge. Notably, the RNN model exhibits early divergence from the ground truth, indicating challenges in capturing evolving dynamics effectively, particularly evident around timestep 5. The LSTM model, while demonstrating a more sustained forecasting performance, begins to show divergence around \(t+20\), suggesting limitations in maintaining accuracy over extended forecast horizons. Both the RC and sequential RC models exhibit comparable performances, preserving dynamics for a more extended range of forecasts, but showing signs of deterioration around \(t+48\). This comprehensive visual analysis sheds light on the strengths and limitations of each model in forecasting SWE dynamics.

\begin{figure}[htbp]
    \centering
    \includegraphics[width=0.8\columnwidth]{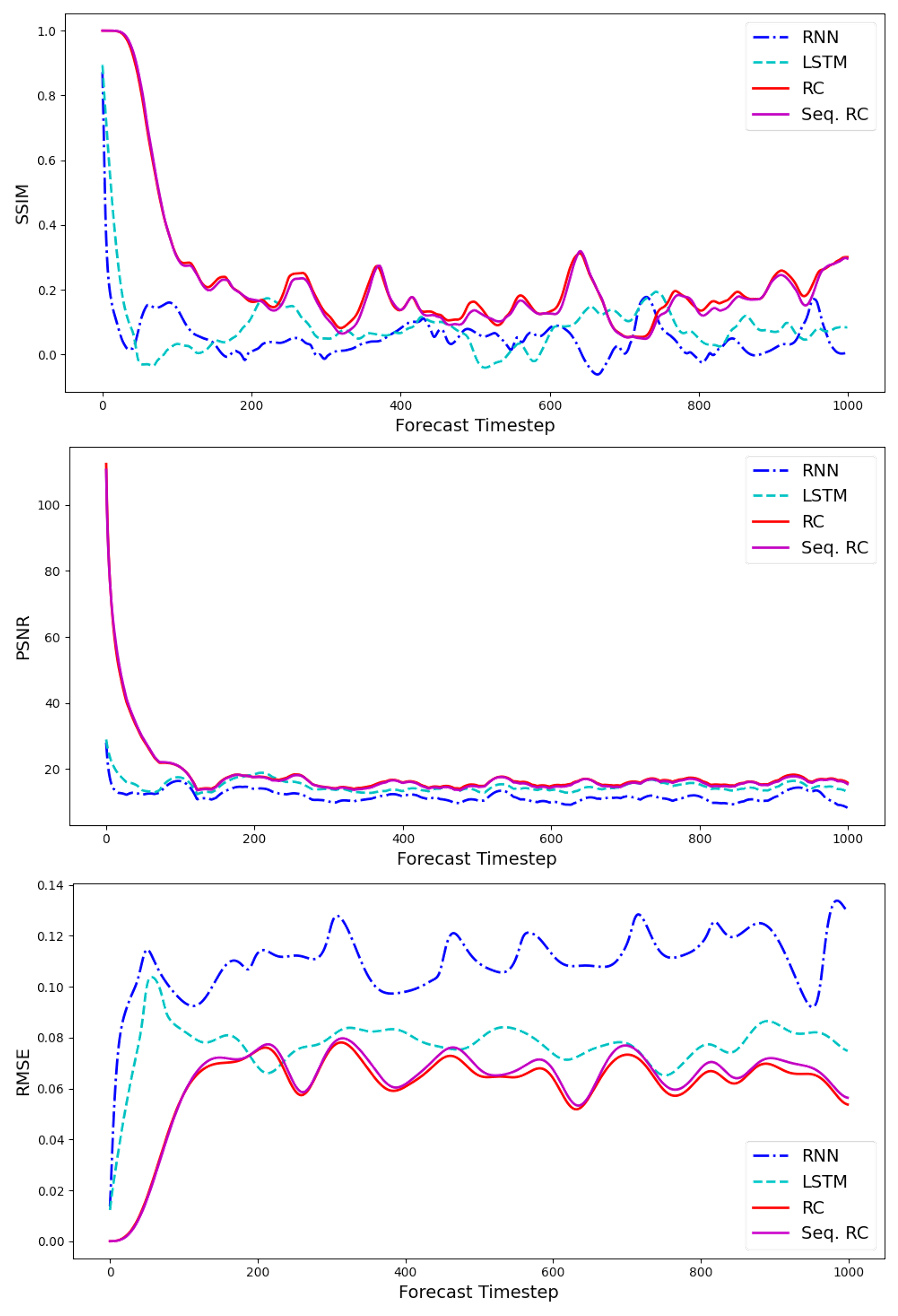} 
    \caption{Quantitative assessment of the SWE dataset using SSIM, PSNR, and RMSE metrics depicted in Figure \ref{fig:swe_graph}. The Y-axis represents metric values, while the X-axis signifies the forecast lead-time, starting from 1. Comparative analysis with Figure \ref{fig:vorticity_graph} reveals the RNN's consistent early deviation from the ground truth, highlighting limited forecasting skills. The LSTM model exhibits a slower decline in metrics, and both RC and sequential RC models demonstrate comparable forecasting skills throughout the lead-time. A decline in SSIM and PSNR is observed for both RC and sequential RC, reaching below the 0.6 SSIM threshold at timesteps 68 and 69. Similar trends are observed in RMSE, with LSTM maintaining relatively better values compared to RNN but inferior to RC and sequential RC.}
    \label{fig:swe_graph}
\end{figure}

In alignment with our approach for Vorticity, we supplemented our visual examination with a quantitative evaluation of the SWE dataset, utilizing metrics such as SSIM, PSNR, and RMSE, as portrayed in Figure \ref{fig:swe_graph}. In these graphical representations, the Y-axis signifies the metric value, while the X-axis denotes the forecast lead-time, initiating from 1. As depicted in Figure \ref{fig:vorticity_graph}, the RNN consistently deviates from the ground truth across all metrics early in the forecast horizon, underscoring its limited forecasting capabilities. The LSTM model demonstrates a more gradual decline from reality across all metrics, though it still encounters a reduction in forecasting performance early in the lead-time. Both the RC and sequential RC models exhibit close performance, showcasing comparable forecasting skills throughout the forecast lead-time. Notably, both RC and sequential RC witness a decline in SSIM and PSNR, falling below the 0.6 SSIM threshold at timesteps 68 and 69, respectively. A similar trend in closeness is observed in RMSE for RC and sequential RC, with LSTM maintaining relatively better RMSE compared to RNN but inferior to RC and sequential RC.

The outcomes detailed in Table \ref{tab:model_train_time_HiDim} provide valuable insights into the performance and computational characteristics of diverse neural network architectures (RNN, LSTM, RC, and sequential RC) under varied conditions. Examining a range of metrics, encompassing trainable parameters, fixed parameters, MegaFLOPS (MFLOPS), GPU memory consumption, and training times, affords a comprehensive perspective on the models' capabilities and efficiencies. The count of trainable parameters serves as an indicator of model complexity, revealing the expected higher capacity of LSTM compared to RNN. Notably, RC and sequential RC exhibit a substantial number of trainable parameters, a consequence of their training as least-squared regression models. This approach leads to significant time savings during training, as reflected in the Training time metric. In the realm of fixed parameters, RC stands out with over 18 times more fixed parameters than sequential RC. Moving to computational complexity, measured by FLOPS, sequential RC demonstrates the lowest FLOPS among the models, indicating a lower computational burden during inference. Memory usage patterns vary, with RNN and LSTM requiring less memory than RC and sequential RC. In the case of RC, this is attributed to extensive reservoir allocation, coupled with input and readout layers. In contrast, sequential RC allocates a major portion of memory to the readout layer and reservoir sequences. However, due to the breakdown of reservoir layers, it registers a smaller memory footprint. Regarding training time, consistent averages are observed between Vorticity and SWE datasets. For datasets comprising 2,000 and 5,000 training samples, RC emerges as the fastest performer. Sequential RC closely trails RC, falling short by only a few seconds in the training duration.

        

\begin{table*}[htbp]
    \centering
    \begin{tabular}{lccccc}
        \toprule
        Metric & Training Samples & RNN & LSTM & RC & Seq. RC \\
        \midrule
        Trainable Parameters (M) & -- & 10.49 & 16.79 & 20.97 & 18.87 \\
        Fixed Parameters (M) & -- & 0.00 & 0.00 & 5.24 & 0.29 \\
        FLOPS (MFLOPS) & -- & 35.65 & 67.10 & 26.21 & 19.17 \\
        Memory (MB) & -- & 80.05 & 128.10 & 192.03 & 146.03 \\
        \midrule
        Training Time (s) & 2,000 & 716.0 & 852.0 & 1.3 & 2.3 \\
        & 5,000 & 1,568.0 & 1,744.0 & 2.4 & 4.9 \\
        \bottomrule
    \end{tabular}
    \caption{Model statistics including trainable and fixed parameters (in millions), MegaFLOPS (MFLOPS), GPU memory consumption for a single forward run, and training time across different dataset sizes for Vorticity and SWE experiments. All experiments conducted on NVIDIA GeForce RTX 3090 with 4 Intel(R) Xeon(R) E5-1660 v4 @ 3.20GHz CPUs.}
    \label{tab:model_train_time_HiDim}
\end{table*}

\section{Conclusion}\label{sec:conclusion}

This study introduced Sequential Reservoir Computing (Sequential RC) — a hierarchical, memory-efficient extension of standard reservoir architectures for spatiotemporal forecasting. By linking multiple small reservoirs in sequence, Sequential RC captures multi-scale temporal features without the computational overhead of backpropagation-based recurrent models.
Across both low-dimensional and high-dimensional chaotic systems, Sequential RC consistently outperformed RNN, LSTM, and conventional RC models. It extended valid forecast horizons by up to 25\%, improved SSIM and PSNR stability over long lead times, and reduced FLOPS and memory requirements by an order of magnitude relative to LSTM. These gains were achieved without increasing the number of trainable parameters and with only marginal additional latency compared to standard RC.


In the low-dimensional domain, we observed that sequential RC exhibited a considerable reduction in the number of operations and demonstrated commendable memory efficiency when compared to RNN, LSTM, and traditional RC models. Our experiments involving three distinct training data lengths (2,000, 5,000, and 10,000 samples) allowed us to discern the impact of varying training dataset sizes on forecasting capabilities. Moreover, our assessment of model efficiency, considering parameters, Floating Point Operations Per Second (FLOPS), and training time, provided valuable insights into the computational efficacy of the models in this context.

Transitioning to high-dimensional datasets, specifically the Vorticity and Shallow Water Equation datasets, our analysis maintained a parallel structure to the low-dimensional exploration. Our objective was to understand how the aforementioned models performed in a more complex data environment. Through a meticulous examination of the influence of training data on forecasting capabilities and the efficiency of models, we sought to unravel patterns and contrasts in computational performance.

The results highlight Sequential RC as a promising framework for real-time, high-dimensional dynamical forecasting—bridging machine learning and physical modeling. Future work will explore adaptive or physics-informed reservoir connections, extensions to turbulent and multi-modal geophysical datasets, hardware implementations on neuromorphic or FPGA systems, potentially porting the architecture to quantum hybrid implementations~\cite{gyurik2025quantum, wudarski2023hybrid, ahmed2025robust}.



\appendix
\section{Appendix A: Historical Window for backpropagation-based models}\label{sec:appendix_a}

\begin{figure}[H]
    \centering
    \includegraphics[width=\columnwidth]{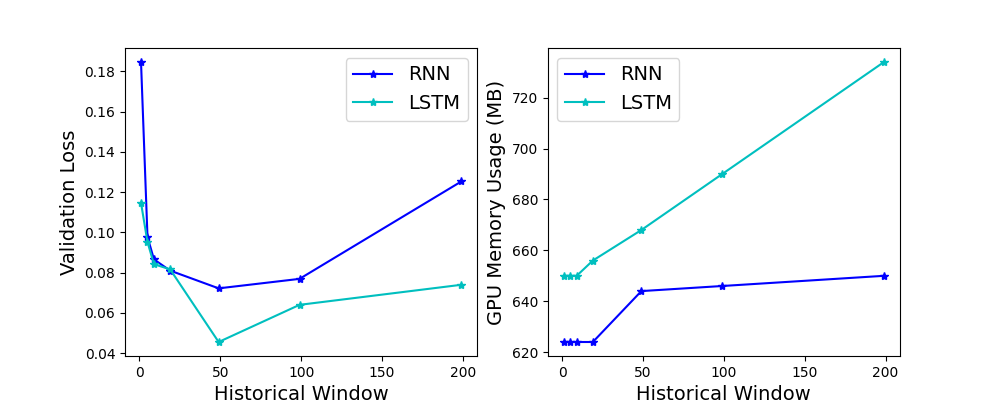}
    \caption{Left: Validation loss reported by RNN (blue) and LSTM (cyan) with different historical window sizes. Notably, both the RNN and LSTM models exhibit a loss minima when using a historical window of 50 input timesteps. Right: GPU memory consumption in megabytes for both the models and intermediate tensors across various historical window sizes.}
    \label{fig:hyperparameter_lookback}
\end{figure}

In Figure \ref{fig:hyperparameter_lookback}, we experimented with the historical window's impact on model performance and model memory consumption. We analyze the impact of varying historical windows on model performance, focusing on both RNN and LSTM architectures. The left subplot showcases the validation loss trends across different historical windows, revealing a noteworthy loss minimum at 50 input timesteps for both model types. This observation suggests an optimal historical context for predictive accuracy that was used for reporting the main results. The right subplot of Figure \ref{fig:hyperparameter_lookback} delves into GPU memory consumption, measured in megabytes, considering both the model itself and intermediate tensors at each historical window. This analysis provides insights into the computational demands associated with different historical windows. The results contribute valuable information for selecting an appropriate historical window size based on a balance between model performance and computational efficiency. It is observed from the figure that LSTM exhibits a significantly steeper increase in GPU demands compared to RNN as the historical window size increases.

\section{Appendix B: Modeling Hyperparameters}\label{sec:appendix_b}

In this section, we present all the hyperparameters used in the models for different simulations in Table \ref{tab:hyperparameters}.


\begin{table*}[htbp]
    \centering
    \begin{tabular}{lp{4cm}lll}
        \toprule
        \textbf{Hyperparameter} & \textbf{Description} & \textbf{Dataset} & \textbf{Model} & \textbf{Value} \\
        \midrule
        Model Precision & Model parameter precision & All & All & Float64 \\
        Data Precision & Dataset precision & All & All & Float64 \\
        Hidden Dimensions & Number of hidden dimensions & Lorenz63 & RNN, LSTM, RC & 256 \\
        Hidden Dimensions & Number of hidden dimensions & Vorticity, SWE & RNN, LSTM, RC & 512 \\
        Sequential Layers & Number of reservoir layers & All & Seq. RC & 8 \\
        Layer Size & Size of each reservoir layer & Lorenz63 & Seq. RC & 32 \\
        Layer Size & Size of each reservoir layer & Vorticity, SWE & Seq. RC & 64 \\
        Learning Rate & Optimizer learning rate & All & RNN, LSTM & 1e-4 \\
        Optimizer & Training optimizer & All & RNN, LSTM & RMSProp \\
        Spectral Radius & Reservoir spectral radius & All & RC, Seq. RC & 1.1 \\
        Leak Rate & Reservoir leak rate & All & RC, Seq. RC & 0.7 \\
        Sparsity & Reservoir sparsity & All & RC, Seq. RC & 0.0 \\
        Historical Window & Length of input sequence & All & All & 49 \\
        \bottomrule
    \end{tabular}
    \caption{List of hyperparameters, their descriptions, applicable datasets, associated models, and values used in all experiments.}
    \label{tab:hyperparameters}
\end{table*}

\section*{Acknowledgments}
USRA authors acknowledge support from Standard Chartered Bank. 
This work used resources available through the USRA Research Institute for Advanced Computer Science (RIACS) and
the National Research Platform (NRP) at the University of California, San Diego. NRP has been developed, and is
supported in part, by funding from National Science Foundation, from awards 1730158, 1540112, 1541349, 1826967,
2112167, 2100237, and 2120019, as well as additional funding from community partners.

\bibliographystyle{unsrt}  
\bibliography{bib}

\end{document}